\documentclass[10pt,journal,compsoc]{IEEEtran}
\usepackage{xspace} 
\ifCLASSOPTIONcompsoc
  \usepackage[nocompress]{cite}
\else
  \usepackage{cite}
\fi

\ifCLASSINFOpdf
  \usepackage[pdftex]{graphicx}
\else
\fi

\usepackage{amsmath}
\usepackage{mathtools}
\usepackage{amssymb}
\usepackage{amsfonts}
\usepackage{soul}
\usepackage{color}
\usepackage{subcaption}
\usepackage{lipsum}
\usepackage{multirow}
\usepackage{booktabs}
\usepackage{cases}
\usepackage{algorithm}
\usepackage{algorithmic}

\begin{document}
\title{DeepForm: Reasoning Large Language Model for Communication System Formulation}
\author{Panlong Wu*~\IEEEmembership{},
        Ting Wang*~\IEEEmembership{},
        \thanks{*These authors contributed to the work equllly and should be regarded as co-first authors.}
        Yifei Zhong~\IEEEmembership{},
        Haoqi Zhang~\IEEEmembership{},
        Zitong Wang~\IEEEmembership{}
        and~Fangxin~Wang,~\IEEEmembership{Member,~IEEE}}
\markboth{IEEE TRANSACTIONS ON MOBILE COMPUTING}
{Shell \MakeLowercase{\textit{et al.}}: Bare Advanced Demo of IEEEtran.cls for IEEE Computer Society Journals}

\IEEEtitleabstractindextext{%
\begin{abstract}
Communication system formulation is critical for advancing 6G and future wireless technologies, yet it remains a complex, expertise-intensive task. While Large Language Models (LLMs) offer potential, existing general-purpose models often lack the specialized domain knowledge, nuanced reasoning capabilities, and access to high-quality, domain-specific training data required for adapting a general LLM into an LLM specially for communication system formulation. To bridge this gap, we introduce DeepForm, the first reasoning LLM specially for automated communication system formulation. We propose the world-first large-scale, open-source dataset meticulously curated for this domain called Communication System Formulation Reasoning Corpus (CSFRC). Our framework employs a two-stage training strategy: first, Supervised Fine-Tuning (SFT) with Chain-of-Thought (CoT) data to distill domain knowledge; second, a novel rule-based Reinforcement Learning (RL) algorithm, C-ReMax based on ReMax, to cultivate advanced modeling capabilities and elicit sophisticated reasoning patterns like self-correction and verification. Extensive experiments demonstrate that our model achieves state-of-the-art performance, significantly outperforming larger proprietary LLMs on diverse senerios. We will release related resources to foster further research in this area after the paper is accepted.
\end{abstract}

}

\maketitle 
\IEEEpeerreviewmaketitle 

\ifCLASSOPTIONcompsoc
\IEEEraisesectionheading{\section{Introduction}\label{sec:introduction}}
\else

\section{Introduction}
\fi

As the cornerstone of modern information society, communication systems are undergoing a paradigm shift from 5G to 6G, playing a pivotal role in advancing emerging domains such as the Internet of Things (IoT), the Industrial Internet, and the Vehicular Internet. These domains rely heavily on ultra-reliable, low-latency, and high-capacity networks to enable seamless connectivity, real-time data exchange, and intelligent decision-making.

The complexity of 6G systems, characterized by their integration of artificial intelligence, terahertz communication, massive machine-type communications, network slicing, and many other cutting-edge technologies, necessitates an accurate modeling framework. Such models provide researchers and engineers with a precise analytical tool to simulate and optimize system behavior under diverse operational conditions. In the advancement of communication technology, communication system formulation has increasingly become a crucial link between theoretical design and practical implementation. By providing a precise framework for understanding and analyzing system behavior, accurate modeling not only enhances the characterization of communication system properties but also lays the groundwork for optimizing performance and facilitating real-world deployment.

However, developing an accurate communication system formulation presents numerous technical challenges. This type of modeling demands a robust understanding of communication principles and often integrates multiple mathematical theories, including information theory, queuing theory, and optimization theory, thereby reflecting significant interdisciplinary complexity.
Moreover, currently, mainstream approaches to communication system formulation exhibit considerable fragmentation across various subfields. Distinct communication system domains—such as integrated sensing and communication (ISAC), massive MIMO, intelligent reflecting surfaces (IRS), and millimeter-wave beamforming—have each evolved their own independent mathematical modeling methodologies. For instance, in ISAC systems, modelers are required to possess dual expertise in both communication signal processing and radar system design. This high degree of specialization has resulted in the emergence of isolated "knowledge silos" within each subfield, hindering cross-disciplinary collaboration and integration. Consequently, overcoming these challenges necessitates not only a deep understanding of individual subfields but also efforts to bridge the gaps between them, fostering a more unified and cohesive approach to communication system formulation.

Recently, large language models (LLMs) such as GPT-4o\cite{hurst2024gpt}, LLaMA\cite{touvron2023llama} and Gemini\cite{team2023gemini}, are driving a paradigm shift in the field of artificial intelligence. These models, with parameter counts of billions, exhibit remarkable capabilities in contextual reasoning and knowledge emergence through self-supervised learning methodology on large-scale data. For instance, GPT-4o, developed by OpenAI not only has complex semantic understanding, multi-turn logical reasoning, and creative content generation but also achieves expert-level performance in structured tasks such as code writing, mathematical proofs, and scientific discoveries. In the field of communication system formulation, LLMs offer new hope for enhancing the efficiency of communication system formulation by leveraging their advanced pre-trained knowledge. 

However, although current LLMs demonstrate outstanding intelligence in open-domain tasks, they still face a severe "capability gap" when applied to communication system formulation. The knowledge of the communication system is characterized by "deep specialization.". General LLMs lack a comprehensive understanding of the communication system, failing to meet the demands of cutting-edge research. Furthermore, existing LLMs are deficient in the deep reasoning abilities specific to communication systems. The complexity of communication system construction underscores the necessity of these deep reasoning capabilities in Communication system formulation. This makes small-scale open-source general LLMs perform poorly in communication system formulation. For large-scale LLMs, they have high deployment costs, for instance, deploying an LLM like DeepSeek R1 requires the use of 8 H100 GPUs, making it prohibitively expensive for many applications. Alternatively, leveraging external LLM APIs for modeling purposes introduces significant privacy and security risks, which hinder the widespread adoption of LLMs in the domain of communication system formulation. 

This reveals the need for adapting a general LLM to Domain-Specific LLMs for communication system formulation, which requires fine-tuning on domain-specific data. However, several challenges arise from this. 

\textbf{Insufficient high-quality communication system formulation training data}. According to the scaling law \cite{kaplan2020scaling}, the quantity of high-quality training data for communication system formulation is essential for the performance of LLMs. However, up to now, there is no large-scale, high-quality datasets in the field of communication system formulation.

\textbf{Deep complexity of communication system formulation task}. Communication system formulation requires a deep understanding of the communication system as well as related mathematical knowledge, such as optimization theory, information theory, etc. This requires injecting communication system reasoning ability into LLM during the training, which is highly difficult.

To fill this gap, in this paper we propose DeepForm, the first reasoning LLM for Communication system formulation. By employing a data-driven approach, our framework significantly reduces the need for expert intervention, thereby minimizing labor costs and shortening the modeling cycle. This ultimately enhances the overall efficiency of the communication system formulation.

We construct the Communication System Formulation Reasoning Corpus (CSFRC), the world's first large-scale dataset specifically designed for complex communication system formulation. Curated from 2015-2025 ArXiv publications (10k+ samples related to communication system formulation), CSFRC addresses critical gaps in existing communication area datasets through two key innovations in two sub dataset. 
The first is \textbf{Reasoning-Centric Supervision set}. The philosophy of the data construction takes deep consideration of the intrinsic property of the communication system formulation area's complexity, making it hard for LLM to learn knowledge directly from the answer, but more suitable to learn through detailed thinking processes.
The second is \textbf{Rule-centric Reinforcement Set}. We develop a sub-dataset which consists of communication system formulation questions and related modeling formulations, specially designed for LLM to get accurate rewards through rule-based RL.

The model training consists of a two-stage training process. 
In the first stage, we conduct the data distillation through supervised fine-tuning on the constructed Chain of Thought (CoT) data to enable the student LLM to have domain knowledge in communication system formulation. 
In the second stage, we propose a rule-based RL algorithm C-ReMax based on ReMax \cite{li2023remax} and is inspired by the amazing performance of rule-based RL in improving general math formulation problem\cite{guo2025deepseek}. The C-Remax algorithm is able to inject complex communication system formulation capability to LLM. The LLM can improve its strategy by answering different types of communication system formulation questions and getting feedback on whether the answer is correct. After being trained by the algorithm, the LLM learn complex reasoning capability in communication system formulation and emerges self correct, verification, back-tracking, and other reasoning behaviors. 

In summary, our contributions are as follows:
\begin{itemize}
 \item To our best knowledge, we construct the world-first large-scale communication system formulation dataset, CSFRC, and will open source it together with relative dataset construction code. 
 
 \item We are the first to propose and open source the reasoning LLM DeepForm for communication system formulation. Our training framework introduces a novel knowledge distillation approach and leverages the potential of rule-based RL in communication system formulation.

\item We conduct extensive experiments, and the results show that the trained model ultimately achieves state-of-the-art performance in communication system formulation.
\end{itemize}

\section{Related Work}

\subsection{Domain Specific LLM in other areas}
The Significant powerful ability of LLMs has attracted many researchers to conduct research on dating a general LLM in other areas. In \cite{zhang-etal-2023-huatuogpt},  the authors  present HuatuoGPT, a medical LLM that combines distilled data from ChatGPT and real-world data from doctors to enhance its performance in medical consultations.  
In \cite{yue2024lawllm}, The authors propose LawLLM, an intelligent legal system with legal reasoning and verifiable knowledge retrieval capability. The system is trained on a high-quality supervised fine-tuning dataset called Law-SFT. The authors also construct a comprehensive legal benchmark, Law-Eval, to evaluate intelligent legal systems from both objective and subjective dimensions. In \cite{chen2023disc}  the authors present DISC-FinLLM, a Chinese financial large language model built using a Multiple Experts Fine-tuning Framework, which enhances general LLMs with capabilities in the finance area. In \cite{bi2024oceangpt}, the authors introduces OCEANGPT, the first large language model specialized in ocean science tasks, present a novel instruction data generation framework called DOINSTRUCT. In \cite{wang2024starwhisper}, it presents the StarWhisper Telescope system, an AI autonomous framework that integrates LLMs with specialized function calls and modular workflows to automate end-to-end astronomical observations. In \cite{luo2024mutaplm}, the authors present MutaPLM, a novel protein language modeling framework that explicitly models protein mutations for enhanced explanation and engineering capabilities through a protein delta network and cross-modal supervision. In \cite{li2024ecomgpt} presents EcomGPT, a large language model fine-tuned on the EcomInstruct dataset, demonstrating superior zero-shot generalization capabilities in e-commerce tasks compared to ChatGPT. In \cite{yang2024mentallama}, the authors introduce MentaLLaMA, the first open-source instruction-following large language model series for interpretable mental health analysis on social media, along with the IMHI dataset, and demonstrate its effectiveness in correctness, explanation quality, and generalizability. In \cite{xiong2023doctorglm}, the authors present DoctorGLM, a healthcare-focused language model fine-tuned from ChatGLM-6B using Chinese medical dialogue datasets and various techniques, achieving cost-effective deployment for medical purposes.

\subsection{Application of LLM in Communication System}
LLMs have recently garnered significant attention for their potential applications in various fields, including communication system. In \cite{tong2025wirelessagent}, the authors propose a wireless agent framework that adapt and enhance LLMs to address the problems in wireless communication system by using prompt engineering, retrieval-augmented generation, and other techniques. The practical applicability is demonstrated in network slicing management. In \cite{10972177}, the authors rethinks generative semantic communication for multi-user systems in 6G and propose the M-GSC framework with a large language model as the shared knowledge base. It highlights three optimization strategies for M-GSC, including extending the LLM-based SKB into a multi-agent system, offloading semantic encoding and decoding, and managing communication and computational resources. A case study demonstrates the preliminary validation of M-GSC's effectiveness in efficient decoding offloading. In \cite{10638533}
The authors propose CommLLM, a novel LLM-enhanced multi-agent system for 6G communications that integrates multi-agent data retrieval, collaborative planning, and evaluation-reflection modules. The system leverages natural language processing and LLMs to overcome challenges in 6G communication tasks by enabling self-learning, self-improvement, and efficient problem-solving. A case study on semantic communication demonstrates CommLLM's effectiveness in autonomously generating and refining communication models to meet specific design requirements. In \cite{10705427}, the authors propose an LLM-based Generative IoT (GIoT) system deployed in a local network setting to address security concerns, which includes a Prompt Management Module, a Postprocessing Module, and a Task-specific Prompts Database to enhance the capacities of open-source LLMs and integrate prompting methods. In \cite{10574890}, a novel LLM-centric Intent Lifecycle (LC) management architecture is proposed for next-generation networks, enabling network configuration and management via natural language. In \cite{10778660}, the authors explore the integration of Large Language Models (LLMs) and graphs in dynamic networking, proposing a novel LLM-enabled graph framework for networking optimization and validating its effectiveness through a UAV networking case study. In \cite{10901300}, the authors  evaluate the effectiveness of LLMs for intrusion detection in IoT networks, proposing a novel LLM-based framework that leverages techniques such as fine-tuning and embedding similarity.  In \cite{10742571}, the authors present a novel framework for trustworthy zero-touch network and service management in 6G by integrating AI for anomaly detection, XAI for root cause analysis, and LLMs for generating user-friendly explanations and implementing corrective actions, demonstrating its efficiency through real-world experiments.

\begin{figure*}[h]
   \centering
   \includegraphics[width=1\textwidth]{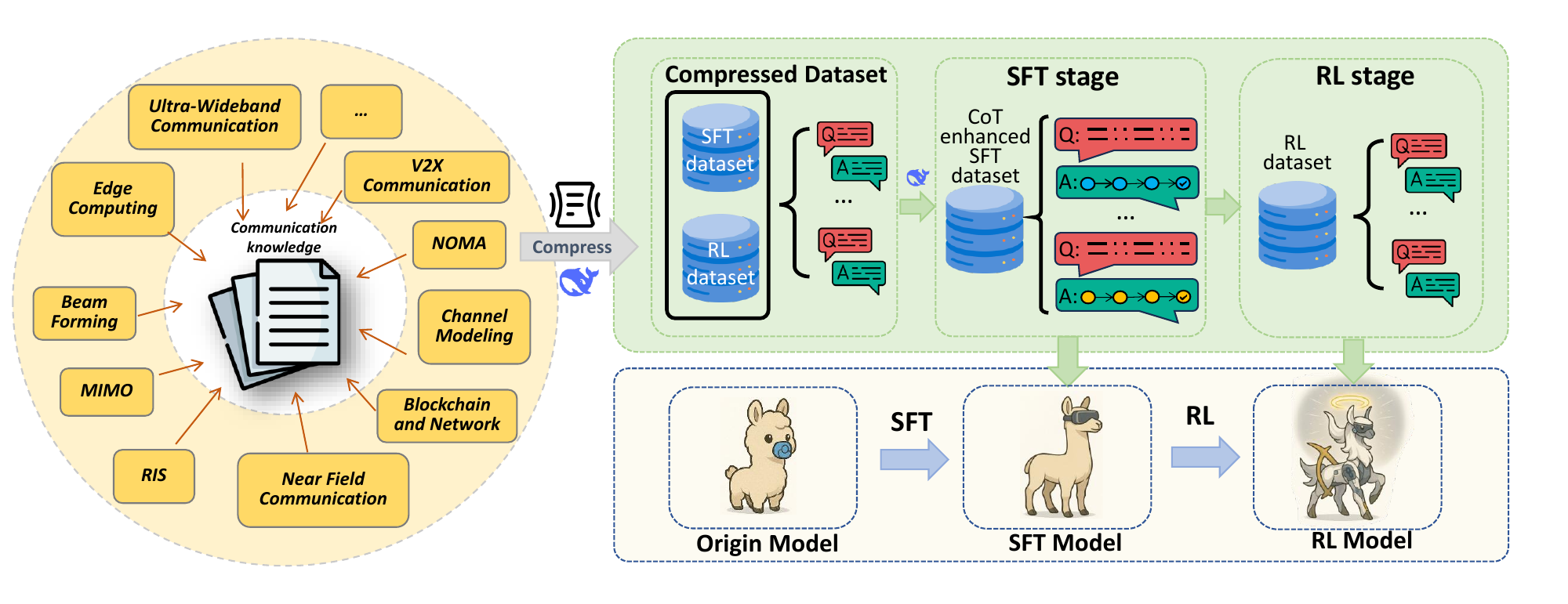}
   \caption{Total Workflow}\label{fig:workflow}
\end{figure*}

\section{Methodology}

\subsection{Data Preprocess Pipeline} 

We systematically retrieved arXiv articles containing wireless communication terminology through the arXiv API. The acquired PDF documents were converted to semantically structured markdown using the \texttt{minerU} library~\cite{wang2024mineruopensourcesolutionprecise}\cite{he2024opendatalab}, employing LaTeX-aware parsing to preserve mathematical notation integrity:
\begin{equation}
    f_{\text{parse}}: \text{PDF}(c_i) \xrightarrow{\texttt{minerU}} \mathcal{M}_i = \langle m_1, m_2, ..., m_K \rangle
\end{equation}
where $\mathcal{M}_i$ represents the document object containing $K$ markdown elements with intact equation blocks ($m_k \in \mathcal{M}_i$).

\begin{figure*}[t]
   \centering
   \includegraphics[width=\textwidth]{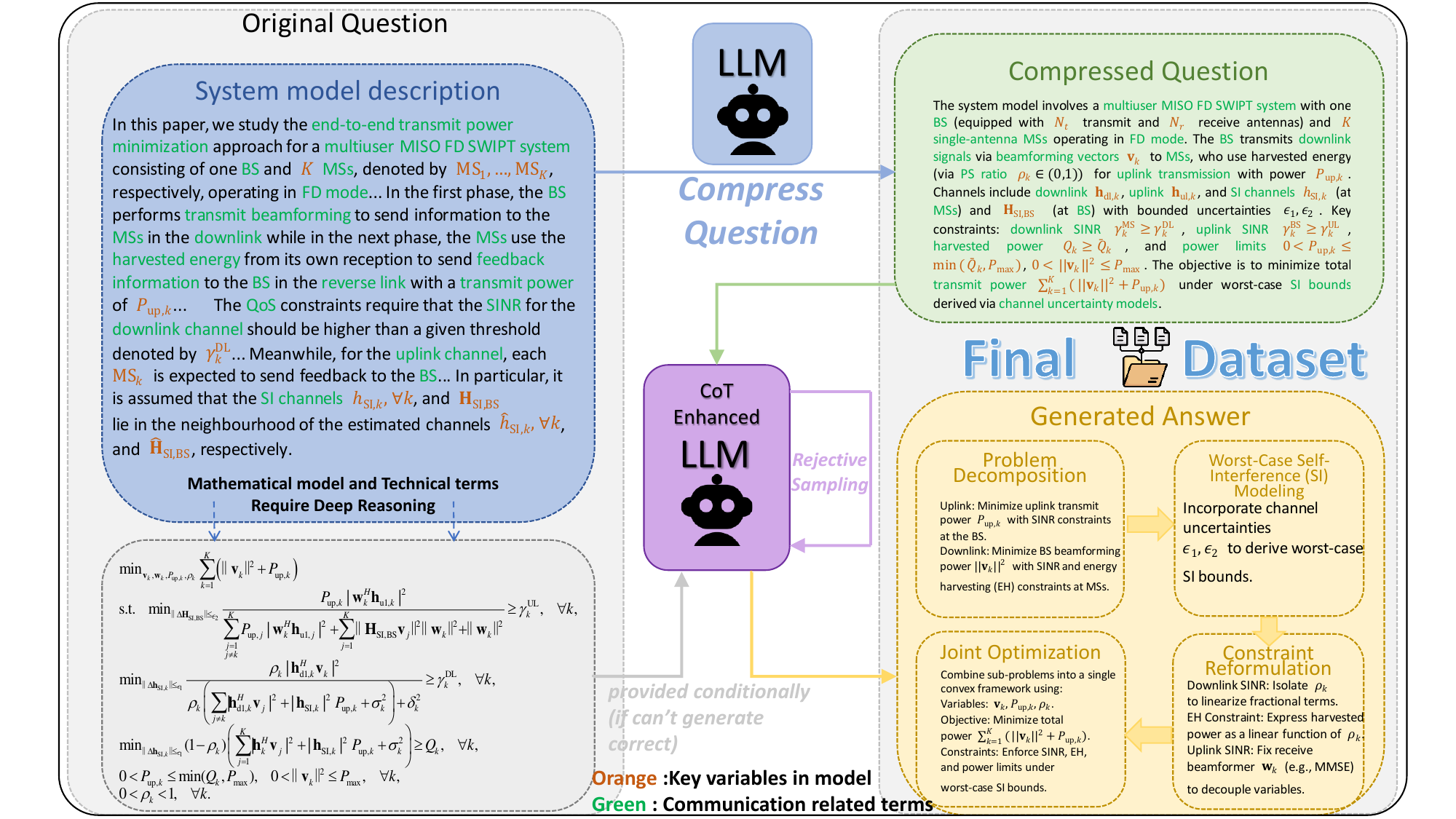}
   \caption{{CoT data detail}}\label{fig:cot}
\end{figure*}

The workflow of the CoT data processing pipeline of a sample is shown in Fig. \ref {fig:cot}, which consists the following part.
\subsubsection{Description-Formulation Pair Extraction}
Technical reasoning pairs were extracted from system modeling sections (specifically targeting subsections labeled "System Model" or "Modeling Framework"). Through regular expression pattern matching, we identified contextual descriptions $d_j$ preceding mathematical formulations $f_j$ in the document flow:
\begin{equation}
    \mathcal{D}_{\text{raw}} = \left\{ (d_j, f_j) \mid d_j \in \mathcal{C}, f_j \in \mathcal{F}, \text{pos}(d_j) < \text{pos}(f_j) \right\}
\end{equation}
where $\mathcal{C}$ denotes contextual descriptions and $\mathcal{F}$ represents formal mathematical expressions.The condition $\text{pos}(d_j) < \text{pos}(f_j)$ indicates that, during the construction of the $D\_raw$ data pairs, all textual descriptions appearing before the position of $f_j$ are collected as the corresponding description $d_j$. In actual data processing, supplementary condition statements following the term ``formulation''— often introduced by words such as ``where'' - are also appended to the description $d_j$. 

\subsubsection{Prompt Compression Optimization}
Leveraging the DeepSeek V3 API \cite{guo2025deepseek} for semantic compression, we applied iterative pruning to description elements:
\begin{equation}
    \hat{d}_j = \text{V3\_compress}(d_j) \quad \text{s.t.} \quad \text{len}(\hat{d}_j) \leq L_{\text{max}}
\end{equation}

where $L_{\text{max}} = 4096$ tokens is chosen to mitigate excessive computational overhead resulting from excessively long prompts during training.
This process yielded the compressed dataset $\mathcal{D}_{\text{comp}} = \{ (\hat{d}_j, f_j) \}$.

\subsubsection{Data Augmentation and labeling} 
The construction of the final dataset employs a two-stage annotation and labeling procedure. \\
\textbf{Candidate Generation.}  
For each compressed prompt $\hat{d}_j$, reasoning paths $\hat{g}_j$ are sampled up to a maximum number of attempts $T_{\text{max}}$:  
\begin{equation}
    \mathcal{G}_j = \left\{ g_j^{(k)} \mid g_j^{(k)} \sim \pi_{\text{R1}}(\cdot \mid \hat{d}_j), \, k=1,...,T_{\text{max}} \right\},
\end{equation}  
 where $\pi_{\text{R1}}$ represents the generator policy utilizing rejection sampling with the DeepSeek R1 API \cite{guo2025deepseek} to evaluate whether a candidate answer aligns with the ground truth. The generation process terminates either when a candidate $g_j^{(k)}$ is confirmed to match the reference answer $f_j$ by the DeepSeek R1 model, or when the maximum number of attempts $T_{\text{max}}$ is reached. If no candidate is validated within $T_{\text{max}}$ trials, the sample proceeds to the Fallback Correction stage (if eligible).

\textbf{Fallback Correction.}  
When the candidate generation fails to produce the correct answer $f_j$, but the highest similarity among generated candidates satisfies  $\text{Sim}(g_j^{(k)}, f_j) \geq \theta$ (Levenshtein similarity\cite{Levenshtein1966}) :  
\begin{equation}
    \tilde{g}_j = \arg\max_{g \in \mathcal{G}_j} \log p_{\text{R1}}(f_j \mid g \oplus \hat{d}_j),
\end{equation}  
where $\oplus$ denotes context concatenation. In such cases, the reasoning path is completed based on the given question $\hat{d}_j$ and the correct answer $f_j$. These samples are labeled as "given-answer CoT samples" in the final dataset to indicate that the reasoning path was reconstructed using the fallback mechanism.  

If $\max_k \text{Sim}(g_j^{(k)}, f_j) < \theta_2$, the sample is discarded without further processing.  

The final dataset is constructed as follows:  
\begin{equation}
    \mathcal{D}_{\text{sft}} = \left\{ (\hat{d}_j, \tilde{g}_j, f_j, \text{label}) \mid j=1,...,M \right\},
\end{equation}  
where $\text{label} \in \{\text{"generated"}, \text{"given-answer"}\}$ indicates whether the reasoning path was directly generated ($\text{"generated"}$) or reconstructed via fallback correction ($\text{"given-answer"}$).

\begin{figure}[h]
   \centering
   \includegraphics[width=0.5\textwidth]{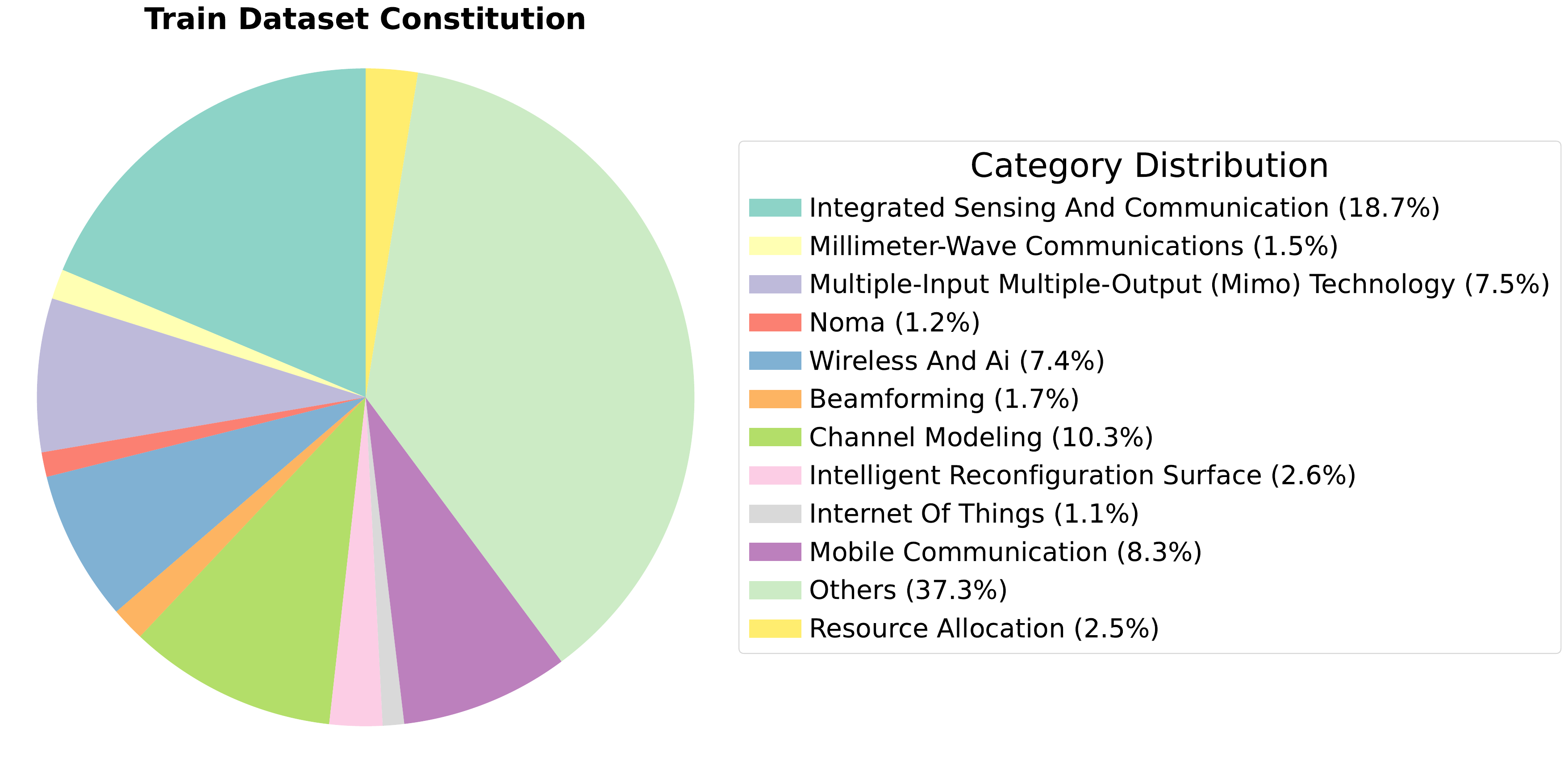}
   \caption{{Training dataset composition}}\label{fig:classification_pie_chart}
\end{figure}

The data constitution is shown in Fig. \ref{fig:classification_pie_chart}, which provides an overview of the diverse composition of the training data used in this study. As illustrated, the dataset encompasses a wide range of categories, reflecting a well-rounded and comprehensive coverage of key topics in communication system formulation. Notably, no single category dominates the dataset entirely, with the largest segment labeled as "Others ($37.3\%$)"—suggesting a balanced distribution across various subdomains. This diversity is further reinforced by the presence of multiple specialized areas such as Integrated Sensing and Communication, Wireless and AI, Channel Modeling, and MIMO Technology, each contributing a meaningful proportion to the overall dataset.

\subsection{Stage One: Learning Complex Reasoning in Communication} 
Direct utilization of LLM to address communication related formulation problems described in natural language often
results in inaccuracies, primarily due to their inability to comprehensively capture implicit information. To address this issue, we enhance our model capability to both define and solve the problems through supervised fine-tuning (SFT).

SFT is a parameter optimization paradigm where a pre-trained large language model adapts to specific downstream tasks through labeled instruction-response pairs. The core components of SFT include the training dataset $\mathcal{D}_{\text{sft}}$ containing description-Chain of Thought (CoT) pairs $\{(x_i, y_i)\}$, and an optimization objective that minimizes the discrepancy between model predictions and ground-truth responses. This structured format enables explicit supervision for both reasoning trace generation and solution verification.

For a general LLM initially lacking domain knowledge in communication system formulation, SFT enforces the injection of the domain knowledge by updating model parameters through gradient-based optimization. To conduct more efficient training, we freeze all the pretrianed parameters and only update LoRA\cite{hu2022lora} parameters. As the update of the LLM has a low-rank nature, we can insert low-rank matrices and only update them during the training. Formally, for a pre-trained weight matrix $W_0$, LoRA approximates the parameter update $\Delta W$ as:  
\begin{equation}  
    \Delta W \!= \!A \!\cdot\! B^T \quad \!\!\!\!\!\text{where}\!\!\!\! \quad A \in \mathbb{R}^{d \times r}, B \in \mathbb{R}^{k \times r} \quad\!\!\!\!\!\! (r \ll \min(d,k))  
\end{equation}  
with $r$ denoting the rank hyperparameter. This decomposition reduces the trainable parameters from $\mathcal{O}(d \cdot k)$ to $\mathcal{O}(r(d + k))$ . 

In our framework, each sample in $\mathcal{D}_{\text{sft}}$ guides the LoRA-augmented model through two sequential phases:
1. Reasoning alignment phase: Given problem description $\hat{d}_j$, the model generates intermediate reasoning tokens $\tilde{g}_j$ under cross-entropy supervision  
2. Solution verification phase: The model then produces the final solution $f_j$ while jointly predicting $\text{label}_j$ through multi-task learning  

The training process optimizes the model's conditional probability distribution $p_\theta(y|x)$ using a composite maximum likelihood estimation objective:  
\begin{equation}  
    \mathcal{L}_{\text{SFT}} = -\mathbb{E}_{(x,y) \sim \mathcal{D}_{\text{sft}}} \left[ \log p_\theta(\tilde{g},f,\text{label} \mid \hat{d}) \right],  
\end{equation}  
where $\theta$ now explicitly incorporates the trainable LoRA parameters $A$ and $B$. This formulation enables gradient updates to focus on low-dimensional subspaces of the original weight matrices, preserving pre-trained knowledge while adapting to task-specific patterns . The optimization alternates between forward pass computation of token-level cross-entropy losses and backward propagation of gradients, typically using AdamW\cite{loshchilov2017decoupled} optimizer with linear learning rate decay.

\subsection{Stage Two: Enhance Complex Reasoning with RL}

\begin{figure*}[h]
   \centering
   \includegraphics[width=\textwidth]{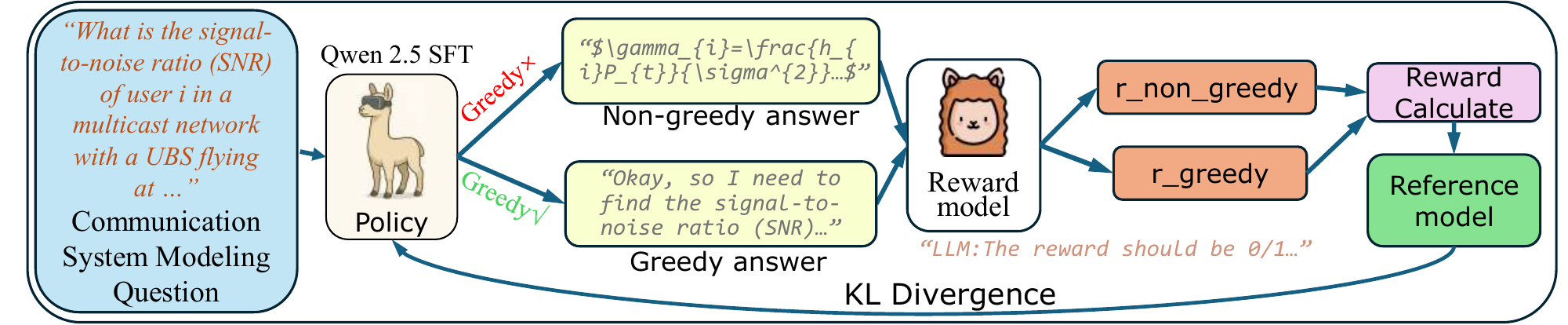}
   \caption{{RL workflow}}\label{fig:template}
\end{figure*}

We propose a rule-based RL algorithm C-Remax based on ReMax\cite{li2023remax}. 

Th RL framework for C-ReMax as a Markov Decision Process is defined by the tuple $\langle \mathcal{S}, \mathcal{A}, P, R, \gamma \rangle$, where:

State Space $\mathcal{S}$: The state $s_t \in \mathcal{S}$ at the time step $t$ is represented by the concatenation of the input communication system formulation question $\mathbf{q} \in \mathcal{V}^*$ and the historical token generated by the LLM $(a_1, ..., a_{t-1})$, where $\mathcal{V}^*$ denotes the vocabulary space of the token sequences. Mathematically, 
\begin{equation}
s_t = \mathbf{q} \oplus (a_1, ..., a_{t-1})    
\end{equation}
where $\oplus$ denotes sequence concatenation.

Action Space $\mathcal{A}$: The action $a_t \in \mathcal{A}$ corresponds to selecting the next token from a discrete vocabulary $\mathcal{V}$, i.e., $\mathcal{A} = \mathcal{V}$. Actions are sampled autoregressively over $T$ time steps to generate answer to the communication system formulation question which can be represented by $\mathbf{y}_{1:T} = (a_1, ..., a_T)$.

Policy $\pi_\theta$: The LLM parameterizes the policy as a conditional probability distribution over tokens:  
  \begin{equation}
  \pi_\theta(a_t | s_t) = P_\theta(a_t | \mathbf{q}, \mathbf{y}_{1:t-1}) \quad \forall t \in [1, T],    
  \end{equation}

where $\theta$ denotes the model parameters and $\mathbf{y}_{1:t-1}$ represents the previously generated tokens.

Reward $R$: 

We use rules to give the reward to the LLM. When the LLM finishes answering the question, it will receive the reward  
If the communication system formulation in the answer of the LLM is correct, we give a positive reward of 1, and if not, we give a zero reward to it. 
In addition to that, we also set a repetition reward to prevent the model to do repetition and make the answer more readable.

The reward is defined as

\begin{equation}
    R(\textbf{a}) = R_a(\textbf{a}) - R_{rp}(\textbf{a})
\end{equation}

where $R_a(\textbf{a})$ represents the accuracy reward and $R_{rp}(\textbf{a})$ represents the repetition reward. The specific definition of accuracy reward is as follows:

\begin{equation}
    R_a(\textbf{a}) = 
    \begin{cases}
    1, & \text{if } \textbf{a} \quad \!\!\!\!\text{equivalent to}\quad \!\!\!\!\textbf{a}_{true} \\
    0, & \text{otherwise}
    \end{cases}
\end{equation}
where $\textbf{a}_{true}$ is the correct modeling formulation and $\textbf{a}$ is the LLM generated answer. 

and the specific definition of repetition reward is as follows:

\begin{equation}
    R_{rp}(\textbf{a}) =  \beta \left(1 - \frac{1}{1 + P(\textbf{a})} \right)
\end{equation}
where
\begin{equation}
    P(\textbf{a}) = \frac{1}{\lambda |\textbf{a}|} 
    \sum_{n=n_{\min}}^{n_{\max}} \sum_{g_n \in \mathcal{G}_n(\textbf{a})} 
    \mathbf{1}[c_\textbf{a}(g_n) > 1] \cdot c_\textbf{a}(g_n)^2
\end{equation}

where $\lambda$ denotes the penalty coefficient. $\{n_{\min}, n_{\max}\}$ denotes the range of n-gram lengths to be counted. $\beta$ denotes the final output upper limit. $\textbf{a}$ denotes the generated token. $\mathcal{G}_n(\textbf{a})$ denotes all n-grams of length $n$ in text $\textbf{a}$. $c_\textbf{a}(g_n)$ denotes the number of occurrences of the n-gram $g_n$ in $\textbf{a}$. $\textbf{1}[ \cdot ]$ is an indicator function, which only counts when the occurrence count is greater than 1.

One critical challenge in communication system formulation arises from its inherent complexity, which induces high gradient variance and hinders LLMs' ability to effectively learn from rule-based feedback signals. In order to reduce the gradient variance, C-ReMax contrasts a stochastic rollout $ \text{seq} \sim \pi_\theta(\cdot|x) $ with a greedy rollout $ \text{seq}_{\text{max}} = \arg\max_{a_{1:T}} \pi_\theta(a_{1:T}|x) $. The reward signal is defined as:
$$
\hat{r}(x, \text{seq}) = r_m(x, \text{seq}) - r_m(x, \text{seq}_{\text{max}}),
$$
where $ r_m(\cdot) $ is a reward model. This baseline is a good approximation of the expected reward $\mathbb{E}_{a_{1:T} \sim \pi_\theta(\cdot | x)} [ r(x, a_{1:T}) ]$, reducing variance while preserving gradient direction.

Also, we incorporates KL divergence constraints to prevent catastrophic forgetting of previous knowledge as 

\begin{equation}
    \mathcal{L}_{\text{KL}} = \sum_{t=1}^T \mathbb{E}_{x,a_{1:t}} \left[ D_{\text{KL}}\left( \pi_\theta(\cdot|x,a_{1:t-1}) \parallel \pi_{\text{REF}}(\cdot|x,a_{1:t-1}) \right) \right].
\end{equation}
where $\pi_{\text{REF}}$ denotes the reference model, 

The final objective combines reward maximization and KL regularization:
\begin{equation}
\mathcal{L}(\theta) = -\mathbb{E}_{x \sim \rho, \text{seq} \sim \pi_\theta} \left[ \log \pi_\theta(\text{seq}|x) \cdot \hat{r}(x, \text{seq}) \right] + \lambda \mathcal{L}_{\text{KL}}(\theta),    
\end{equation}

where $\lambda$ balances reward and regularization. Gradients are computed via:
\begin{equation}
\nabla_\theta \mathcal{L}(\theta) = -\mathbb{E}_{x,\text{seq}} \left[ \nabla_\theta \log \pi_\theta(\text{seq}|x) \cdot \hat{r}(x, \text{seq}) \right] + \lambda \nabla_\theta \mathcal{L}_{\text{KL}}(\theta).    
\end{equation}

\subsubsection{Dataset construction for RL}

\section{Experiments}

\subsection{Experimental Setup}
The experiment was conducted on two servers: one with 8 A100 GPUs and another with 8 A6000 GPUs, both running Ubuntu 20.04. All training was performed on the CSFRC dataset. The first stage utilized 5k data samples, while the second stage used 1.2k data samples.

Using the proposed method, we trained our models based on Qwen2.5-7B-Instruct \cite{yang2025qwen3}. In Stage 1, the models were fine-tuned on the SFT dataset for 5 epochs with a learning rate of $5 \times 10^{-6}$ and a batch size of 32. We employed LoRA tuning with a rank of 256 and used cosine learning with a warmup ratio of 0.05. In Stage 2, we applied reinforcement learning (RL) with a learning rate of $2 \times 10^{-6}$, a batch size of 64, and a total of 5 training epochs. The KL divergence was set to 0.001.

\begin{figure*}[t]
   \centering
   \begin{subfigure}[t]{0.33\linewidth}
      \centering
      \includegraphics[width=\linewidth]{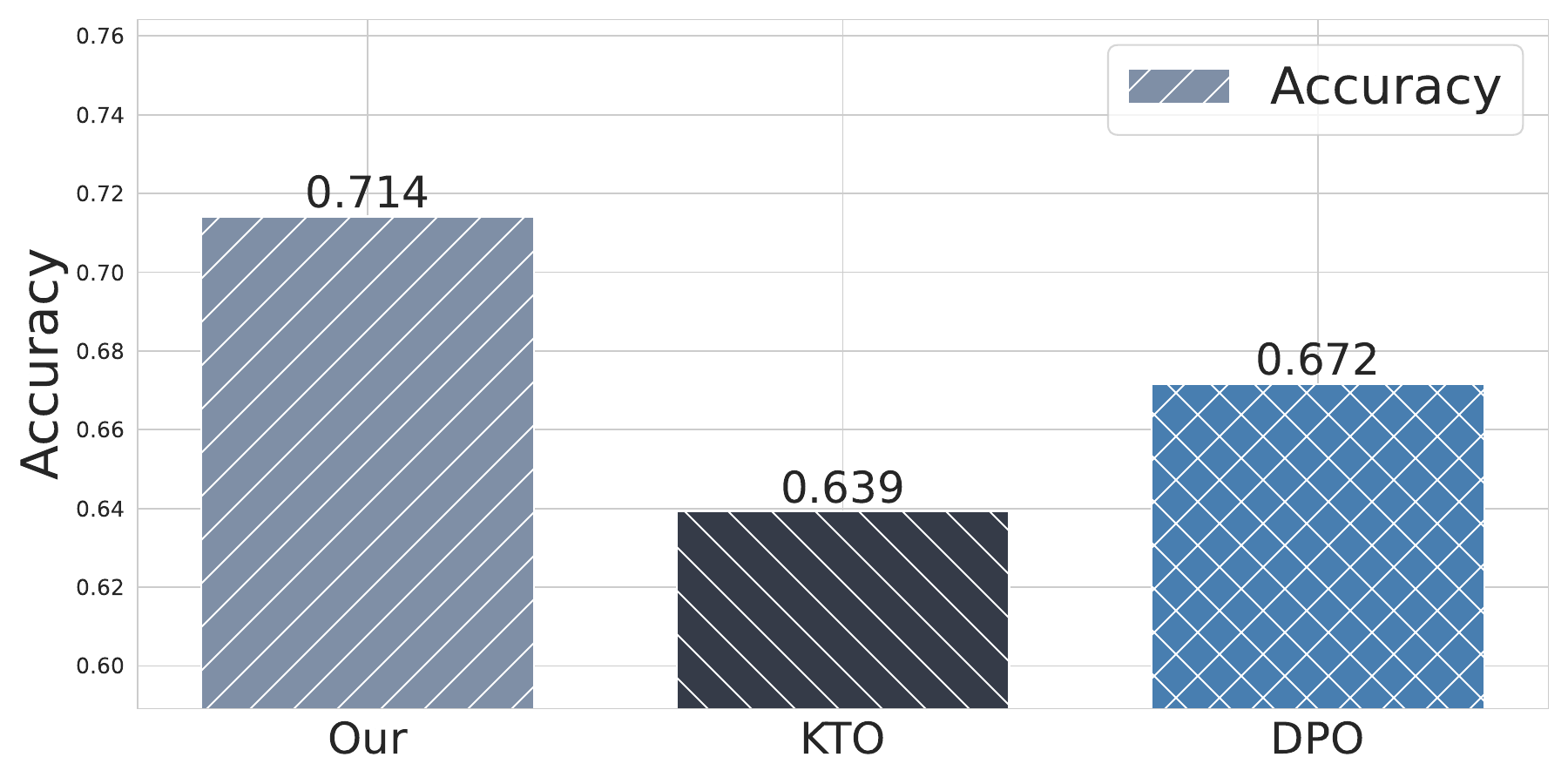}
      \caption{Performance comparison of different RL algorithms}
      \label{fig:template1}
   \end{subfigure}
   \hfill
   \begin{subfigure}[t]{0.33\linewidth}
      \centering
      \includegraphics[width=\linewidth]{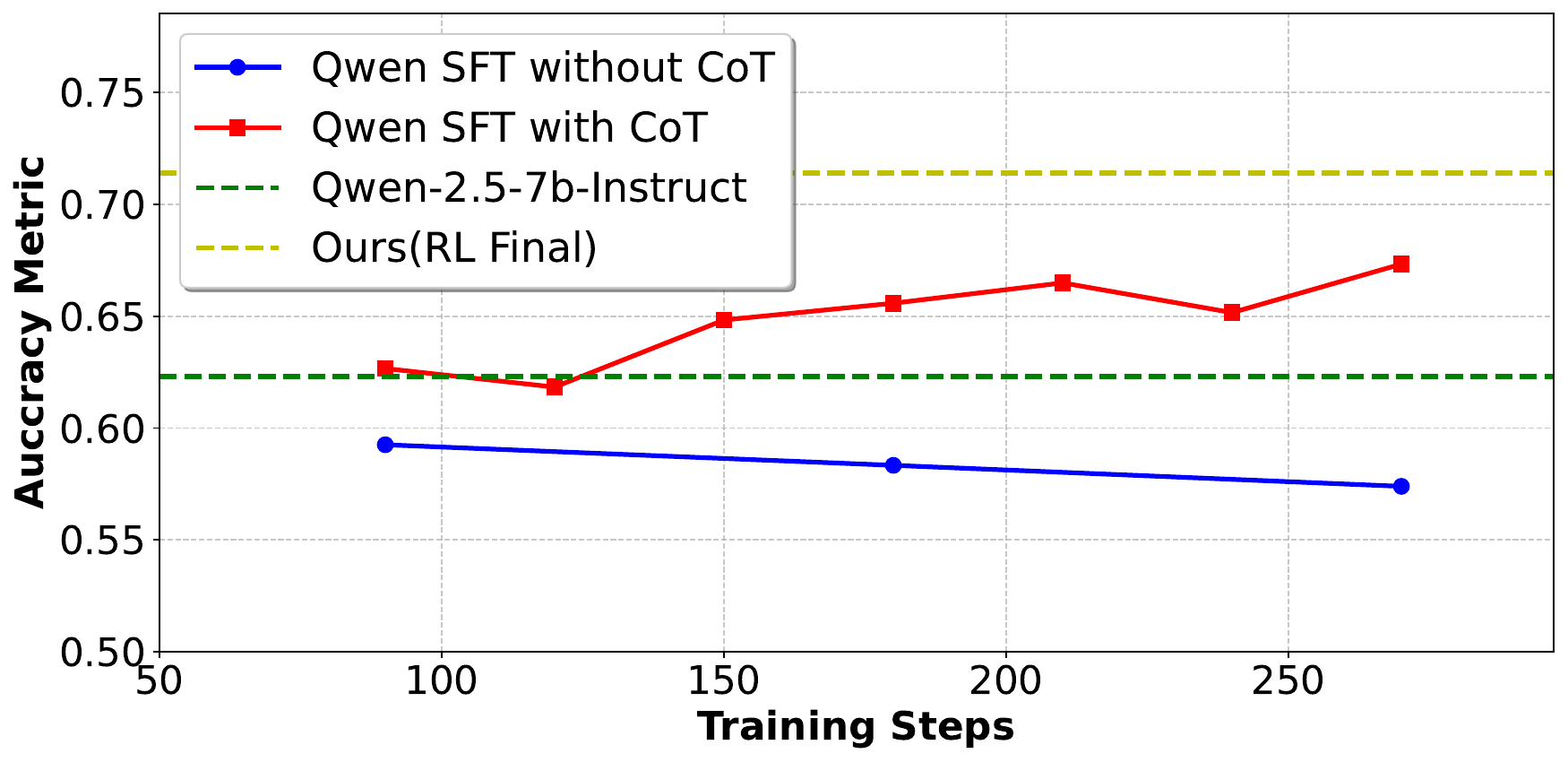}
      \caption{Performance comparison of SFT W/O CoT training data}
      \label{fig:template2}
   \end{subfigure}
   \hfill
   \begin{subfigure}[t]{0.33\linewidth}
      \centering
      \includegraphics[width=\linewidth]{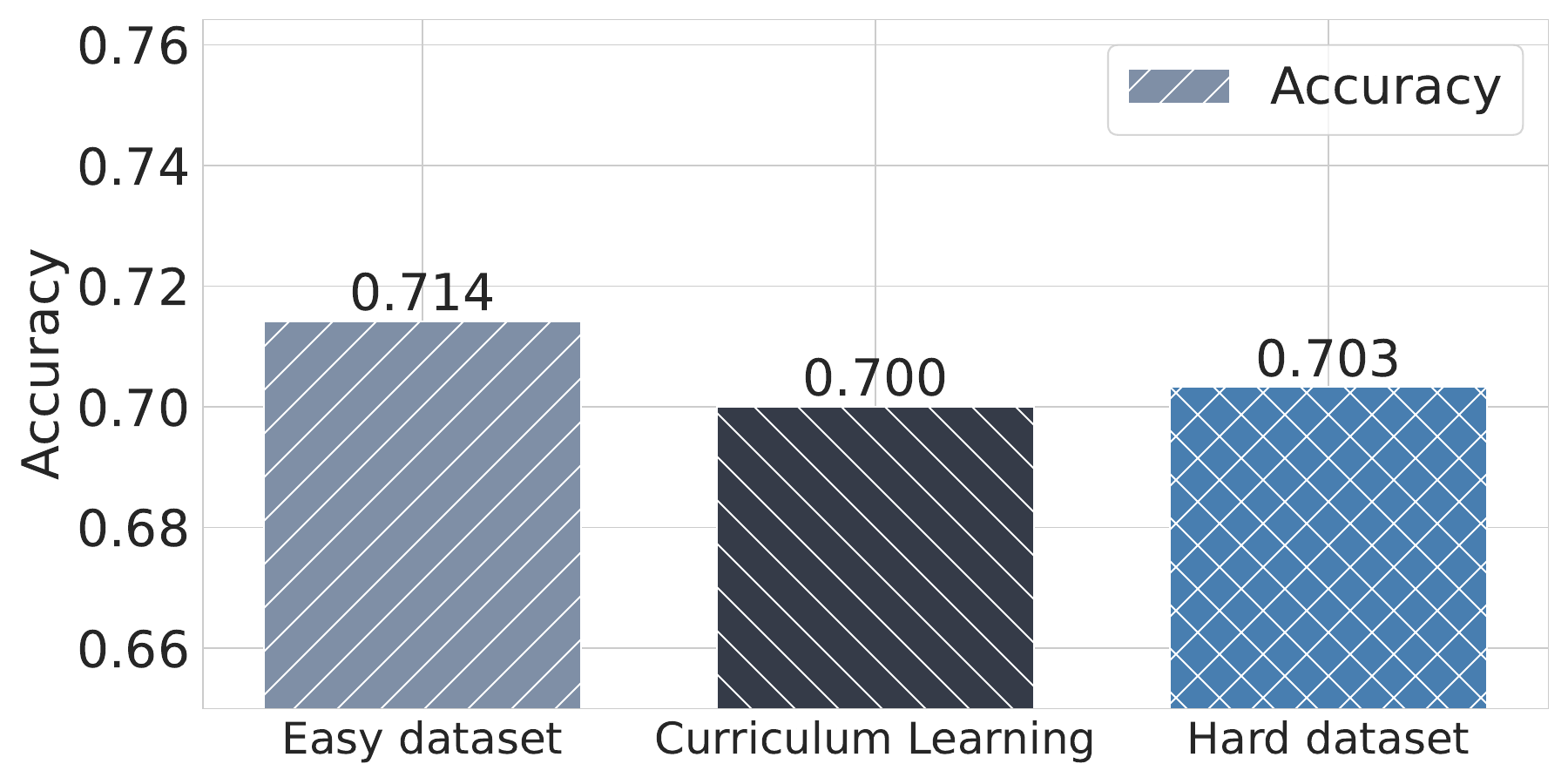}
      \caption{Performance comparison of RL with different datasets}
      \label{fig:diff_dataset}
   \end{subfigure}
\end{figure*}

\subsection{Experimental Results}

\textbf{Performance comparison with other LLMs}

\begin{figure}[t]
   \centering
   \includegraphics[width=0.5\textwidth]{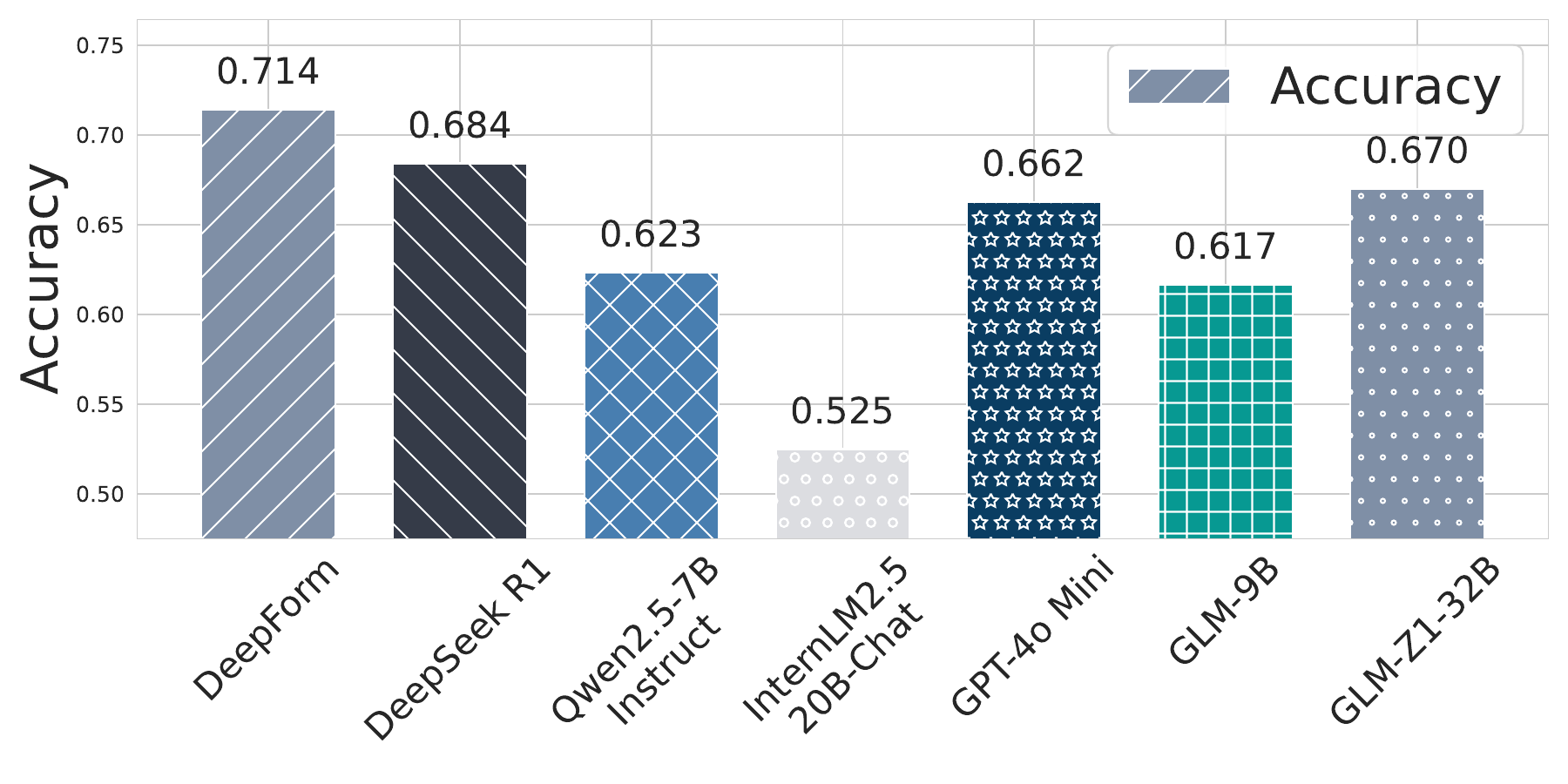}
   \caption{{LLM comparison}}\label{fig:main_results}
\end{figure}

Fig. \ref{fig:main_results} shows the accuracy of different LLMs on the test dataset. It presents a comparative analysis of model performance through a bar chart, where the x-axis lists 6 distinct LLMs and the y-axis quantifies their accuracy scores. The models evaluated include "DeepForm(Ours)," "DeepSeek R1," "Qwen2.5-7B Instruct," "InternLM2.5 20B-Chat," "GPT-40 Mini," and "GLM-9B." 
From the result we can find that DeepForm achieves the highest accuracy among all the LLMs with a model size of only 7B. It significantly outperform the second-highest LLM which is DeepSeek R1 that has a parameter size of 671B which is 94 times larger than our model.

\textbf{Performance comparison of different RL algorithms}\\
We compare our algorithm with several different RL algorithm to further verify the effectiveness of our algorithm. More specifically, we compare with the DPO\cite{rafailov2023direct} and KTO\cite{ethayarajh2024kto}.

As these algorithms require example answers to enable the LLM to learn, we construct the dataset in the following ways.

\textbf{Data construction of DPO.} 
For DPO, we generate pairwise preferences using the DeepSeek R1 to compare reasoning paths:
$$
\mathcal{D}_{\text{DPO}} = \left\{ (\hat{d}_j, g_j^{\text{chosen}}, g_j^{\text{rejected}}) \mid j=1,...,N \right\},
$$
where $g_j^{\text{chosen}}$ and $g_j^{\text{rejected}}$ are two reasoning paths for the same question $\hat{d}_j$, annotated by DeepSeek R1  and Doubao1.5pro (developed by Bytedance) as the preferred and dispreferred outputs, respectively. 

\textbf{Data construction of KTO.}
For KTO, it only need a single generated reasoning path for each sample and each sample requires a label. we generate the reasoning path and annotate single-sample preferences using the DeepSeek R1:
$$
\mathcal{D}_{\text{KTO}} = \left\{ (\hat{d}_j, g_j, s_j) \mid j=1,...,K \right\},
$$
where $s_j \in \{0,1\}$ is a binary label indicating whether the reasoning path $g_j$ is deemed acceptable ($s_j=1$) or not ($s_j=0$).

\begin{figure}[h]
   \centering
   \includegraphics[width=0.5\textwidth]{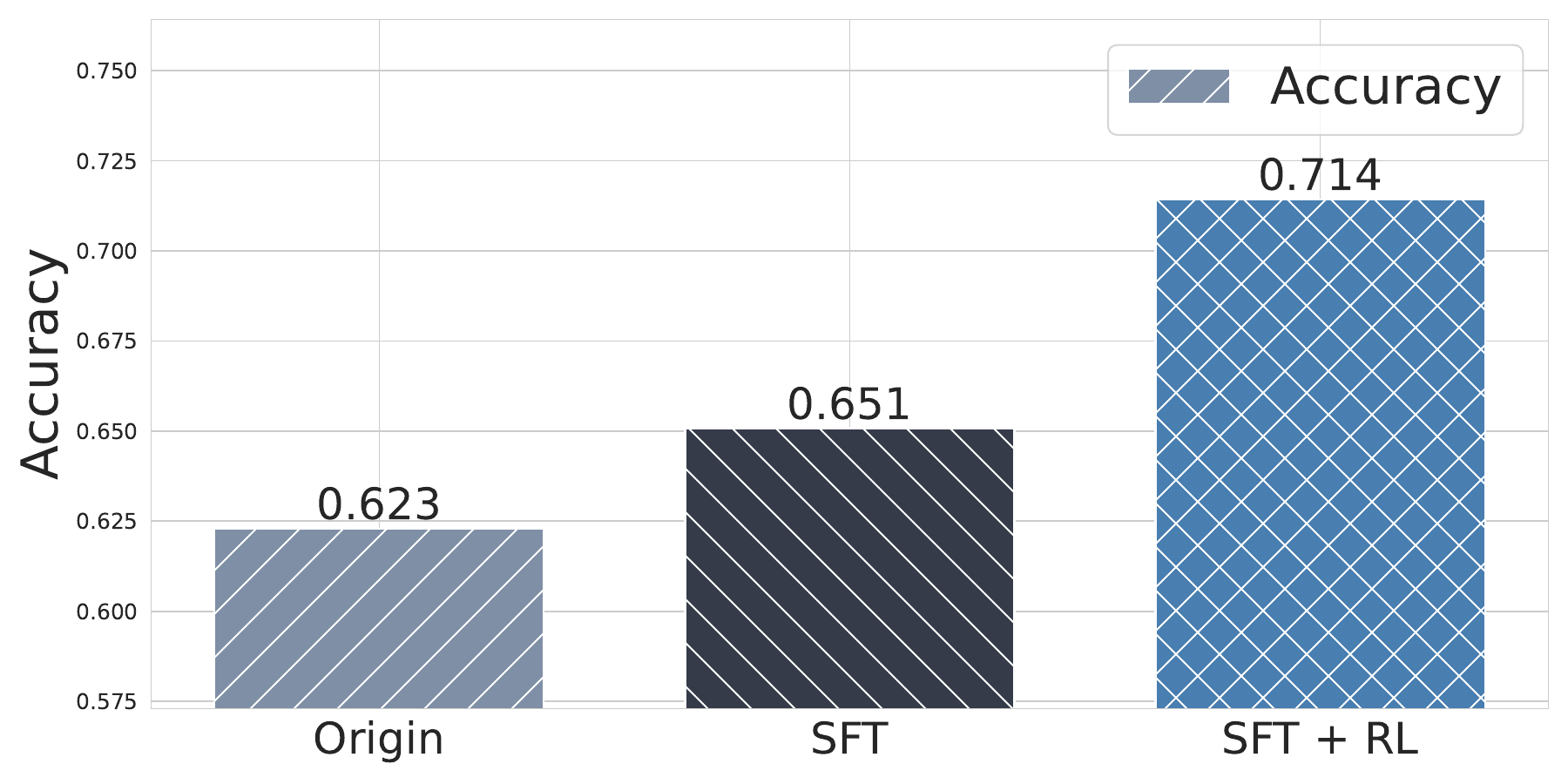}
   \caption{{Ablation Study}}\label{fig:ablation}
   
\end{figure}

\textbf{Ablation Study.} 
We conducted an ablation study on the model to analyze the impact of SFT and RL. The results are shown in Fig. \ref{fig:ablation}. The LLM achieves an accuracy of $65.1\%$ after the SFT stage and further increases to $71.4\%$ after the RL stage.
By observing the results, we can find that both the SFT stage and the RL stage play a critical role in improving the ability of the LLM in communication system formulation. 

\textbf{Performance comparison of SFT W/O CoT training data.}

In this section, we conduct supervised fine-tuning (SFT) experiments on the Qwen 2.5-7B-Instuct model using two distinct datasets to validate the necessity of the proposed chain-of-thought (CoT) data augmentation methodology. For the non-CoT-enhanced dataset, we employ $\mathcal{D}_{\text{comp}} = \{ (\hat{d}_j, f_j) \}$, which contains compressed descriptions $d_j$ preceding mathematical formulations $f_j$. Conversely, the CoT-enhanced dataset $\mathcal{D}_{\text{cot}} = \left\{ (\hat{d}_j, \tilde{g}_j, f_j, \text{label}) \mid j=1,...,M \right\}$ is utilized for comparative analysis to verify the effectiveness of the CoT augmentation approach. Both datasets are used to fine-tune the Qwen 2.5-7B-Instuct model for 270 training steps under identical hyperparameter configurations. Subsequently,  average accuracy evaluations are performed on the same validation dataset using the training prompts. The experimental results are summarized in ~\ref{fig:template2}.

As shown in ~\ref{fig:template2}, the direct use of the non-CoT-enhanced dataset not only fails to improve the model's answer accuracy, but also somewhat diminishes its reasoning ability. We believe that for complex modeling problems, the reasoning process is a crucial component. Merely providing questions and answers can mislead the model's learning of modeling to some extent.
These results demonstrate that sample quality plays a critical role in communication domain modeling, and the CoT data augmentation proves essential for achieving effective knowledge distillation in our dataset configuration.

\textbf{Performance comparison with different difficulty level RL datasets}.
We compare the performance of the LLM after RL using datasets of varying difficulty. Two datasets were constructed: an easy dataset and a hard dataset. We trained the LLMs in three different ways: exclusively on the easy dataset, exclusively on the hard dataset, and through curriculum learning by first training on the easy dataset and then on the hard dataset. The results, shown in Fig. \ref{fig:diff_dataset}, indicate that the LLM trained only on the easy dataset achieved the highest accuracy. The LLMs trained on the hard dataset and through curriculum learning showed similar, but lower, accuracy levels. This discrepancy is likely due to the hard dataset causing instability during RL, as the LLM rarely receives positive rewards because it often fails to provide correct answers, making it difficult for the model to learn from feedback.

\begin{figure}[h]
   \centering
   \includegraphics[width=0.5\textwidth]{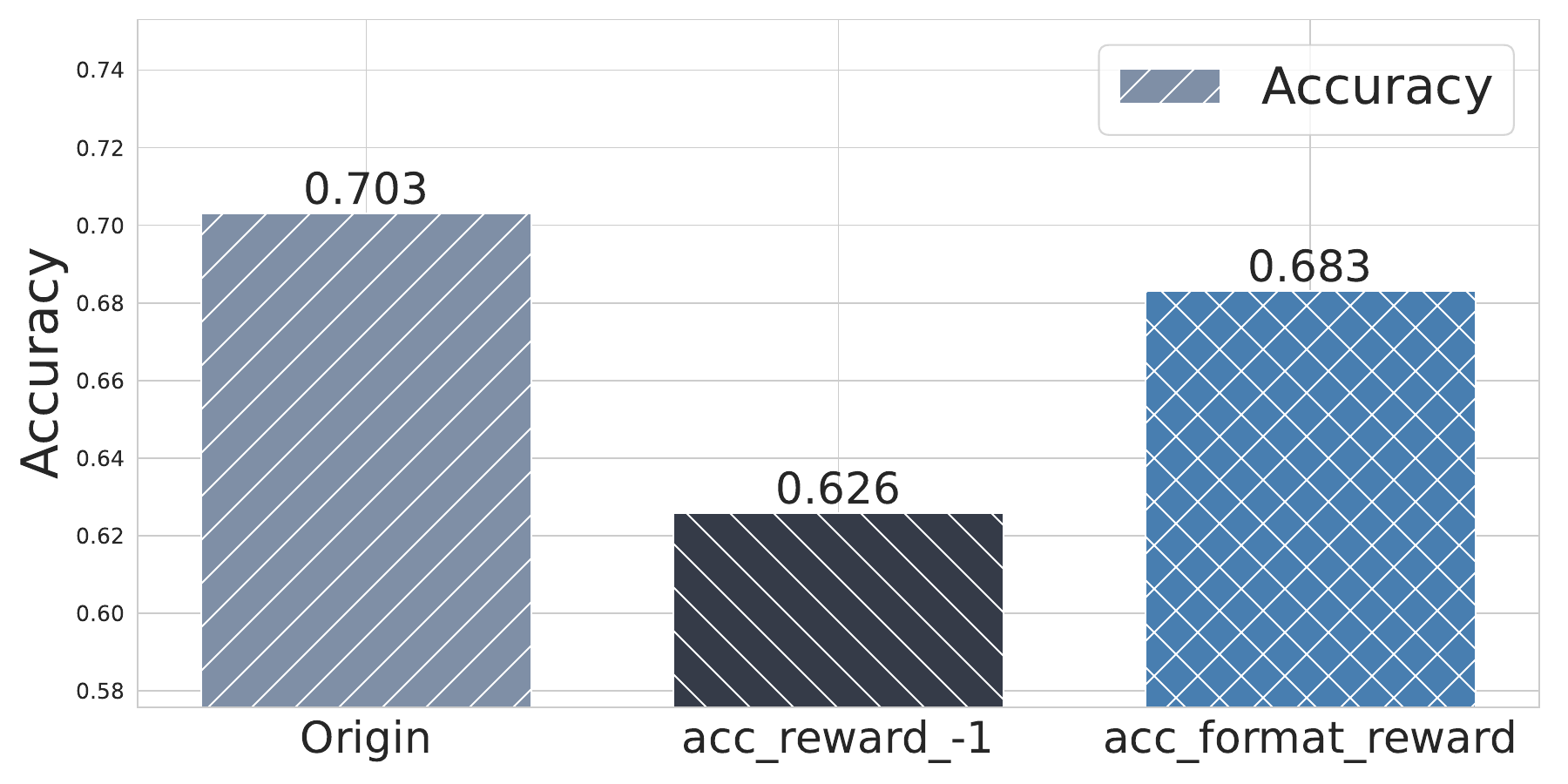}
   \caption{{Comparison of performance with different reward settings}}\label{fig:diff_reward}
\end{figure}

\textbf{Performance comparison of different reward settings.}
In this part, we compare the performance of the LLM with different reward settings in RL. The experiments 

We conduct training on different reward penalties when the LLM provides the wrong answer. We experiment under the cases that a reward of "$-1$" is given to the LLM and a reward of "$0$" is given to the LLM if it provides the wrong answer.
The two reward settings are as follows.

\begin{subequations}
\begin{align}
    R_{\textbf{a}}(\textbf{a}) &= 
    \begin{cases}
    1, & \text{if } \textbf{a} = \textbf{a}_{\text{true}} \\
    0, & \text{otherwise}
    \end{cases} \label{eq:binary_reward} \\
    R_{\textbf{a}}(\textbf{a}) &= 
    \begin{cases}
    1, & \text{if } \textbf{a} = \textbf{a}_{\text{true}} \\
    -1, & \text{otherwise}
    \end{cases} \label{eq:penalty1_reward} 
\end{align}
\end{subequations}

In addition to that, we also test the effect of the existence of the format reward. To address both semantic accuracy and format compliance, we propose a composite reward function combining:

- \textbf{Accuracy reward}: Measures correctness of the response.

- \textbf{Format reward}: Ensures adherence to format requirement.

The format reward is defined as:
\begin{equation}
    R_{f}(\textbf{a}) = 
    \begin{cases}
    1,  & \text{if format is correct} \\
    -0.1,  & \text{otherwise}.
    \end{cases}
\end{equation}
and the combined reward is defined as:
\begin{equation}
    R(\textbf{a}) = R_{f}(\textbf{a}) + R_{a}(\textbf{a})
\end{equation}

The performance comparison under different reward settings is illustrated in Fig. \ref{fig:diff_reward}. The results indicate that a change in the accuracy reward setting can lead to an accuracy drop of (7.7\%). This may be attributed to the LLM frequently receiving negative rewards, which can cause instability during training. Additionally, the inclusion of an external format reward in the reward function results in a (2\%) decrease in performance. This decline could be due to the LLM's reduced exploration to avoid negative format rewards.

\section{Conclusion}
In this work, we present a comprehensive framework for adapting LLMs to the domain of communication system formulation, addressing critical challenges in insufficient high-quality communication system formulation training data and the deep complexity of the communication system formulation task. Our contributions are threefold. Fisrt, we introduce CSFRC, the world’s first large-scale open-source dataset for communication system formulation, and will open-source it for researchers to do further research. Second, we are the first to 
develop a novel two-stage LLM training framework specially for communication system formulation, which contains a CoT data distillation stage and a rule based RL stage. We train the world-first reasoning LLM named DeepForm in communication system formulation and have done extensive experiments to prove the capability of DeepForm.

\IEEEdisplaynontitleabstractindextext

%

\ifCLASSOPTIONcompsoc

\ifCLASSOPTIONcaptionsoff
  \newpage
\fi



%


\bibliography{reference.bib}

%








\bibliographystyle{IEEEtran}

\end{document}